# Sparse distributed localised gradient fused features of objects

Swathikiran Sudhakaran[a], Alex Pappachen James[a,b,*]

[a] Enview R&D Labs
[b] School of Engineering, Nazarbayev University

Abstract

The sparse, hierarchical and modular processing of natural signals are characteristics that relate to the ability of humans to recognise objects with high accuracy. In this paper, we report a sparse feature processing and encoding method targeted at improving the recognition performance of automated object recognition system. Randomly distributed selection of localised gradient enhanced features followed by the application of aggregate functions represents a modular and hierarchical approach to detect the object features. These object features, in combination with minimum distance classifier, results in object recognition system accuracies of 93% using ALOI, 92% using COIL-100 databases and 69% using PASCAL visual object challenge 2007 database, respectively. Robustness of object recognition performance is tested for variations in noise, object scaling and object shifts. Finally, a comparison with 8 existing object recognition methods indicated an improvement in recognition accuracy of 10% in ALOI, 8% in case of COIL-100 databases and 10% in PASCAL visual object challenge 2007 database.

Keywords: Object recognition, Object features, Sparse features, Feature fusion, Hierarchy, Modularity, Brain inspired systems

## 1. Introduction

Humans can identify objects from naturally recorded images even under a diverse range of natural variability such as noise, pose changes and missing features, with high accuracy and speed. This robustness and accuracy of recognition is attributed to the structural and functional organisation of brain processing. It is understood that hierarchical structure of the brain processing, and the ability to integrate the information in a modular form attributes to this innate robustness. In the past, this idea from neuroscience has been explored by the computer scientist to successfully develop techniques such as hierarchical temporal networks [22], memory networks [25] and cognitive algorithms [23] for useful pattern recognition applications. In this paper, we inspire from the hierarchical processing, modularity and additional constraint of sparsity in developing a set of features useful for automated object recognition. We demonstrate that incorporating these ideas in the automated object recognition task can result in improved recognition accuracies.

The ability of the brain to process the information in a sparse manner allows the freedom of recognising objects even when the feature descriptions are partial and incomplete. This property is tightly linked not just to this functional aspect, rather also to the structural organisation of the brain. While the brain is constituted of billions of interconnected neurons, the information about a specific object is represented as an activation of a small number of neurons. Depending on the strength of the signal i.e. how frequently the object is observed by the human, the neuron responds in a distributed manner across the neuron network to encode and store the object features. These features are sparse in representation and offer several advantages [34]. The primary advantage is the ability to identify and store discriminative information with limited number of features. Since neuronal firing is a charge governed operation, the overall energy consumption is reduced by minimising the number of neurons used for the representation of object. In computing terms, a reduced complexity of the data resulting from localised feature processing with sparsity constraints and hierarchy often results in lower complexity in implementing real-time systems that often appreciate a data driven scalable parallel computing solution.

[*]Tel: +7 (7172) 709133
Email address: apj@ieee.org (Alex Pappachen James)
URL: www.biomircosystems.info/apj (Alex Pappachen James)

In terms of signal processing literature, the sparse representation or sparse coding consists of generating a basis set for a signal that can be represented as linear combination of the elements from the set [40]. The main feature of this sparse representation is that only a small number of features are required for the representation of signals. This is analogous to the small number of active neurons within the large scale neural network that corresponds to discriminant object features. The disadvantage of this sparse coding technique is that these are signal dependant, i.e., the basis set generated for one set of signals cannot be used for others. Another major disadvantage associated with the sparse coding techniques is the computational complexity possessed by these techniques. The basis set is computed as an optimization problem requiring complex computations [26, 4, 28, 38] and is seen to have application in sparse representation of images [33]. Currently, the sparse coding of images is applied in many fields such as image denoising [30, 18], super-resolution [51, 50], face recognition [49], background modelling [12, 16], and image classification [29, 52].

Sparse representation of objects inspires from a number of biological vision theories [8, 27, 36, 47] and has shown to result in improved computational efficiency [3, 32]. The sparse representations reduces the dimensionality of the input data thereby simplifying the processing of the data. Although, reduction of the redundant information can lead to reduced feature dimensionality, the philosophy of sparse processing does not warrant reduction in data itself, instead it is more reflective of the preservation of useful information. The term "useful" here is strictly dependent on the application itself. For example, for object compression, it is the useful information that should reveal and reconstruct the details of the object, while for object recognition the identifying and preserving the distinct and discriminative features of the object become most useful.

In this paper, we inspire from the sparse processing ability of human brain, in developing a method for hierarchical and modular processing of features that involve sparse pixel selection, feature grouping, feature encoding and feature fusion. The resulting features are used for object recognition application, in a view to increase the intra-class similarities and to increase the inter-class differences between the object representations, which is reflected through the reported recognition accuracies. We show that these steps results in improved recognition accuracy of the object recognition task, and supports our view to inspire from the vision processing of biological systems in object recognition problems.

2. Background

The image features represented as pixels can be represented as a linear regression problem with the unknown feature values projected onto a lower dimensional space $y = \Phi x + n$, where $x \in \Re^N$ is the feature vector, $y \in \Re^M$ is the measurement vector with $M < N$, $\Phi \in \Re^{M \times N}$ is the measurement matrix, and $n \in \Re^M$ is the additive noise. The measurement matrix is formed from the random selection of feature values from the images under consideration. The extraction of $x$ from $y$ in general is ill-posed as $M < N$ and $\Phi$ would have non-trivial null space. This would mean that there are infinitely many solutions for identifying the true $x$ [1, 2, 5, 6, 7]. This problem is of importance when the question at hand is signal recovery, and it not very much a serious issue in object recognition problem. However, the discriminative information if preserved by the signal recovery mechanisms by ignoring the redundant information can be useful for object recognition. A direct sparse representation such that $x = \Psi \alpha$ in basis $\Psi \in \Re^{N \times N}$ with $K \ll N$ coefficients of $\alpha$ is sufficient to approximate the signal $x$. This can be seen as a basis pursuit problem. The determination of any $\Phi$ on $x$ is the process of encoding and the process of recovering the $x$ from $y$ is referred to as decoding. Both these operations have a wide range of applications, nonetheless, in this paper, the attempt is on developing an encoding scheme. The development of any decoding scheme is ignored in this paper as its practically not required for an object recognition problem.

Among the leading examples of encoding and decoding are compressive sensing techniques [6, 7], that is used in a variety of signal processing applications. In addition, there exists several algorithms [10] for solving such basis pursuit problems by treating this as a subset of convex optimisation problem [11]. Some of the applications include signal component separation, missing data estimation, deconvolution, denoising and signal representation. Although, sparsity is widely used in signal processing with a view to find a generic model for representing object features, much of these thoughts correlate with the sparsity constraints critical for learning biologically plausible models. A common approach to impose such constraint is by inhibiting the less dominant responses, and using only the most dominant responses for feature encoding [34].

We incorporate the idea of biological systems to ignore the components of the signals that are least dominant to enforce the sparsity criteria. The localised nature of features is the another peculiarity with object features, as they are geometrically and spatially connected. However, these connectedness does not always ensure a distinctiveness in recorded images, as even minor variability between the test and training images, results in a statistically significant differences between them. As opposed to the approaches that impose geometric constraints [21, 9], we ignore the geometric constraints and form a randomised selection of features to create group of features. The idea of feature groups can be observed in some of the prominent object recognition methods [14].

Hierarchical processing is another major component of the human brain and visual system[15], which we try to accommodate in the feature encoding process. As opposed to the idea of hierarchical processing as observed in [42], we propose to implement the multi-level processing through the selection of the dominant features along the feature bit planes followed by digital to analog conversion. The digital to analog conversion can be seen as a subset of aggregate feature fusion method implemented using weighted aggregate operator function [17].

3. Proposed method

3.1. Sparse Distributed Fused Features of Objects

The proposed feature extraction process inspires from hierarchy, modular processing and sparsity constraints that reflect typical characteristics of brain and cognitive processing. Pixels in images form the primary unit of representation of object. These units when arranged spatially across two dimensions form a region of object and several such regions when geometrically placed in a visually perceptive manner leads to image representation of an object. Object information is encoded in a sparse manner within the physically and functionally localised neuron modules. This process of encoding features incorporates sparseness constraints that enable modular and hierarchical processing of features. The proposed approach creates the pixel groups by randomly selecting the pixels from the original image across its bit planes. After the feature groups are processed in parallel using a localised sparse selection, aggregation and binarization, the bit planes are fused to obtain encoded features.

---

**Algorithm 1** Feature extraction using sparse distributed representation

---

Input: Image(I) of dimension m × n
Output: Feature vector subset ($F^*$) of length C = k × m × n/W

1: Convert the image I to P bit planes
2: for p=0 to P − 1 do
3:     Select $p^{th}$ bit plane
4:     for c=1 to C do
5:         Choose W number of binary pixels $b^*$ from I randomly
6:         $B(c) = \sum_{l=1}^{W} b_l^*$
7:     end for
8:     Divide the feature vector $B_p$ into groups of X number of $B^*$ feature cells
9:     $B_p = [\{B_p(1), B_p(2)..B_p(C/X)\}, ..., \{B_p(C - C/X), B_p(C - 1)..B_p(C)\}]$
10:    for every element in a group in $B_p$ do
11:        $F_p^*(c) = \begin{cases} 1, & B^*(c) = \max(B^*) \\ 0, & \text{otherwise} \end{cases}$
12:    end for
13: end for
14: for c=1 to C do
15:     $F^*(c) = \sum_{p=1}^{P} 2^{p-1} F_p^*(c)$
16: end for

---

The list of steps involved in the proposed system is shown in Fig. 1 and formally described in Algorithm 1. We consider a quantised model of pixels for our algorithm, such that the integers when converted to the binary representation can be used to form image bit planes. As an example, consider an image with 8 bit pixel representation that can be used to create 8 bit planes or 8 binary images. The image is divided into several regions, and feature cells are formed from each of the distinct regions by randomly selecting the bits from it. Each element of the feature cell is formed by repeated and sparse selection of bits from random locations in the region followed by the summation of such selected bits.

The process of obtaining the feature cell $F^*(c)$ requires the stages outlined in Fig. 1. The image I is binarized along its P bit planes to from binary images $I_b$. Given $R_b \in I_b$ is a region in the image, the binary pixel $b^* \in R_b$ is randomly selected W number of times, and aggregated to form the binary feature $B(c) = \sum_{l=1}^{W} b_l^*$. The binary feature vector B is partitioned into several subsets $B^*$. The binary feature element $F_p^*(c)$ in plane $p \in [0, P]$ is determined as:

$$F_p^*(c) = \begin{cases} 1, & B^*(c) = \max(B^*) \\ 0, & \text{otherwise} \end{cases} \qquad (1)$$

The element $F^*(c)$ in feature cell can be determined as the summation of binary weighted product typically used for binary to continuous data conversion:

$$F^*(c) = \sum_{p=1}^{P} 2^{p-1} F_p^*(c) \qquad (2)$$

Suppose there are W pixels in the selected for generating the $F^*$ feature subset, several such cells form the feature vector F of the image. The length of the feature vector is given by $(k \times m \times n)/W$, m and n are the height and width of the image. The value of k determines whether there is any overlap of pixels among cells, and is referred to as the degree of overlap. If the value of k is equal to one, then the pixels are equally shared among the cells. If the value of k is less than one, then some of the pixels in the image will not be selected during the pixel selection step. In contrast, if the value of k is greater than one, some pixels will get selected by multiple cells. Fig 1(b) shows that the pixel at the location (5,1) was selected by both cell 1 and cell 2. The selected inputs are combined using aggregation operator to result in feature values of a cell. After computing the feature values of all the cell elements, we arrange the entire cells into several groups, such that each group contains X number of cells. These cells are computed in parallel and may be considered as the first level of modular processing. Inside each group, the cell with the highest value will be changed to 1 and the remaining cell's value will be changed to 0. This binary operation signify the identification of most dominant value, and is inspired from the idea that neurons in the brain respond to the most dominant input responses that is often referred to as winner-take-all principle [35]. We repeat this operation for all the groups, and for all the bit planes. The resulting set of binary feature groups across the 8 bit planes represent the most dominant feature locations, and we consider this as a second stage of modular processing in analogy with limited number of fired neurons across the neuron network. In the final stage, the final feature vector is obtained by converting the 8 bit planes to integer form. This fusion process is essentially a weighted aggregate operation, and results in a global encoded feature vector of the object. The multiple stages of localised operations on pixel groups indicate hierarchical processing and the encoded features reflect sparse representation of the object.

3.2. Gradient Enhanced Features

Human eye has the ability to detect the spatial change between the neighbourhood of pixels. The sensory neurons are responsible for detecting the changes as it responds to inter-pixel intensity changes across image space. This idea has been widely used in image processing community for developing first and second order gradient filters that calculate gradient features from the summation of weighted differences between the neighbourhood pixels across the entire image. To incorporate this aspect into our feature encoding scheme, we extend the idea of sparse selection of pixel bits and aggregative operators on gradients of pixel intensities.

In the proposed method, three different variants of features F are extracted from the images to capture different types of object features. The first feature vector is extracted directly from the pixel intensity features as outlined in Section 3.1, while the remaining two feature vectors are extracted from the gradient of the image. The gradient images are formed using the Sobel [44] and Prewitt [39] operator, and the vector of the gradient magnitude and gradient angles are used for generating the feature vector. Let I be an image of dimension m × n, then the gradient image G is given by:

$$G(i, j) = I * w = \sum_k \sum_l I(i-k, j-l)(w(k, l)) \quad (3)$$

where, ∗ represents the convolution operation, and w is the kernel matrix that defines the Sobel [44] and Prewitt [39] operators. Let $G_x$ and $G_y$ be the gradient images in the horizontal and vertical directions respectively. The gradient magnitude image |G| is formed as:

$$|G(i, j)| = \sqrt{G_x(i, j)^2 + G_y(i, j)^2} \quad (4)$$

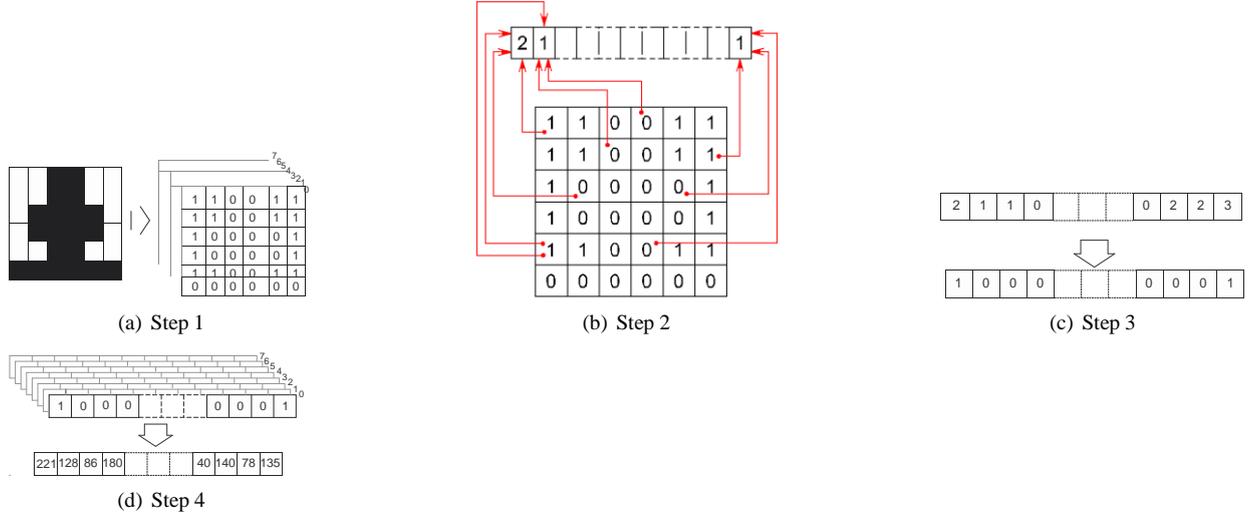

(a) Step 1

(b) Step 2

(c) Step 3

(d) Step 4

Figure 1: The proposed feature extraction and encoding process is illustrated in (a) -(d). In (a), the first step, the input image I is converted to bit planes $I_b$. The second step (b), selects the pixels randomly from each bit plane region $R_b$ and adds the bit values to form cell feature B(c). In the third step (c) the cells are divided into groups. As an example four cells are selected to form groups. Then the value of the cell that has a maximum value is replaced with 1 and remaining cells within the group with 0 to form $F^*_p(c)$. Finally, in (d) i.e. step 4, the binary feature vectors are combined using the weighted addition operation to obtain the final feature vector F.

The feature vector is generated from this gradient magnitude image using the sparse distributed representation technique illustrated in Fig 1. The second feature vector is generated from the gradient direction. The gradient direction $G_\theta$ is formulated as:

$$G_\theta(i, j) = \tan^{-1} \frac{G_y(i, j)}{G_x(i, j)} \quad (5)$$

The resulting gradient direction image will be having values in the range [$-90^0, 90^0$]. To avoid 2's complement computations resulting from negative numbers, the range of values is normalised to [$0^0, 180^0$] by adding $90^0$ to remove the negative values. The feature vector will be then generated from this gradient direction image.

The vector consisting of pixel values I, gradient magnitude |G| and gradient direction $G_\theta$ form the primary input to the Algorithm 1 instead of just the pixel values I. This results in a feature vector F that is three times bigger in dimensions, however, incorporates the spatial change features improving the robustness of feature representation.

3.3. Algorithm Parameters

The number of times pixels are selected (window size) W, number of elements in feature subset (cell size) X and the degree of overlap factor k are the major parameters of the proposed object feature extraction method. The length of the feature vector is determined by these parameters. The window size W determines the number of pixels used to determine the value of element within each feature subset. The value of W will be greater than one and is limited by the number of pixels present in the image. Larger the value of W, more will be the number of pixel values that are encoded into a single feature vector cell. A larger value of W ensures a wider coverage of variations in the pixels in the object and ensure robustness the feature value calculation at the expense of higher computational cost. Feature subset or cell size (X) determines the number of inputs to be given to the winner take all network. X number of feature vector cells are selected and value of the cell with the highest value will be changed to one, while values of remaining cells will be changed to zero. This process is continued for the next set of X cells until all the feature vector values are changed to zero or one. As the cell size(X) increases, the number of pixels applied to the winner take all method also increases and as a result, the effectiveness of the algorithm in combating the local and global variations will get reduced. The degree of overlap (k) determines whether a single pixel is used to more than one number of cells. If the value of k is greater than one, then it means a single pixel is used by more than one feature vector cell. For k less than one, all the pixel values will not be selected for feature representation and can result in loose of discriminative information for forming the object features. The optimal parameter values selected for the experiments explained in Section 4 are W=16, X=4, k=2. The explanations for the same are given in Section 4.3.

## 4. Experiments and Results

### 4.1. Experimental Setup

The developed sparse distributed localised gradient fused features along with a minimum distance nearest neighbour classifier [13] form the modules of the object recognition experiment. In minimum distance classification, the distance between the test feature vector and each of the training feature vectors is computed and the pair with the least distance is selected as the most similar. In this paper, nearest neigbour classification is performed with three different metrics namely, Cityblock distance, Euclidean distance and Shepard's similarity measure [43]. Equations (6), (7) and (8) show the cityblock distance, Euclidean distance and Shepard's similarity function between two feature vectors $F_{train}$ and $F_{test}$. The pair with the least distance value is selected as the similar pair in cityblock distance and Euclidean distance, while for Shepard's similarity measure, the pair with the highest value is selected as the most similar pair. This is because Shepard's similarity measure represents the perceptive degree of similarity between two sequences rather than the distance as is given by cityblock and Euclidean distances.

$$d = \sum_{i}^{X} |F_{train}(i) - F_{test}(i)| \tag{6}$$

$$d = \sum_{i}^{X} (F_{train}(i) - F_{test}(i))^2 \tag{7}$$

$$d = \sum_{i}^{X} e^{-|F_{train}(i) - F_{test}(i)|} \tag{8}$$

In order to check the performance of the proposed method, object recognition experiments was performed on three different standard databases namely, Amsterdam library of object images (ALOI) [24], Columbia object image library (COIL-100) [31] and PASCAL visual object challenge 2007 database [20]. ALOI database contain images of 1000 different objects while the COIL-100 database contain images of 100 objects. The ALOI database contains images with dimension 192 × 144 pixels where as the dimension of the images in the COIL-100 database is 128 × 128 pixels. In both the databases, for each object, 72 different images are present, each taken at a different rotation angle at steps of $5^0$. The training set was formed by choosing images taken at $45^0$ intervals. i.e., for each object class, eight images are used as training set. The remaining images are used as the test set. The PASCAL visual object challenge (PVOC) 2007 dataset contains images of 20 different objects. The training set consists of 5011 images and the test set consist of 4952 images. From each image, the required object is cropped using the given annotations and the dimension of the cropped object is changed to a standard size of 192 × 192 pixels. The recognition accuracy is calculated as the percentage of images correctly classified. Figure 2 shows sample images from the ALOI and COIL-100 database. In all the experiments mentioned in this section, the feature vector consists of pixel values, gradient magnitude and gradient direction with parameter values W=16, X=4 and k=2.

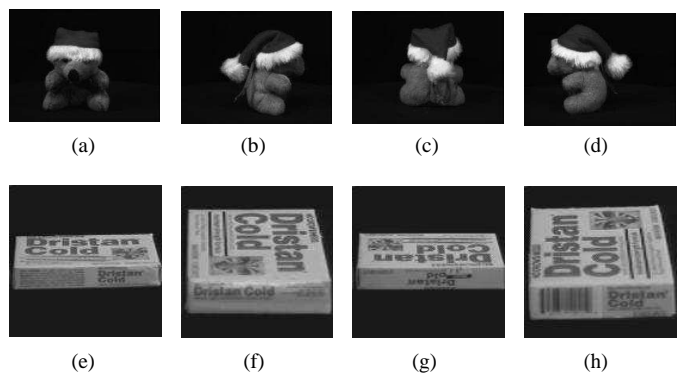

Figure 2: Sample images from the ALOI and COIL-100 databases. (a), (b), (c) and (d) shows an object with $0^0$, $90^0$, $180^0$ and $270^0$ rotations, respectively from the ALOI database. (e), (f), (g) and (h) shows an object with $0^0$, $90^0$, $180^0$ and $270^0$ rotations, respectively from the COIL-100 database

4.2. Recognition Performance

The performance of the system is evaluated in terms of recognition accuracy. The recognition accuracy is computed as the ratio of number of images correctly classified to the total number of images used for classification. Seven different feature vectors were used for object recognition, each formed using different combinations of the three feature vectors explained in Section 3.2. The recognition accuracies for each of these feature vectors are listed in Table 1. In Table 1, the feature pix represents the feature vector generated from the image, feature mag represents the feature vector generated from the gradient magnitude image, feature dir represents the feature vector generated from the gradient direction image. Features pixmag, pixdir, magdir and pixmagdir represents the different combinations of the pix, mag and dir features.

Table 1: The recognition accuracy for various features and distance metrics with parameters W=16, X=4, k=2

| Feature | Distance measure | Recognition accuracy (%) | |
|---|---|---|---|
| | | Prewitt mask | Sobel mask |
| pixmagdir | City block | 91.43 | 91.95 |
| | Euclidean | 90.27 | 91.17 |
| | Shepard | 93.18 | 92.90 |
| pixmag | City block | 91.04 | 91.41 |
| | Euclidean | 88.21 | 89.74 |
| | Shepard | 89.57 | 90.49 |
| pixdir | City block | 89.28 | 89.31 |
| | Euclidean | 87.81 | 87.84 |
| | Shepard | 91.17 | 90.96 |
| magdir | City block | 90.51 | 91.57 |
| | Euclidean | 89.68 | 90.59 |
| | Shepard | 87.79 | 85.27 |
| pix | City block | 87.40 | |
| | Euclidean | 85.32 | |
| | Shepard | 90.77 | |
| mag | City block | 89.09 | 90.39 |
| | Euclidean | 87.94 | 89.15 |
| | Shepard | 84.43 | 79.90 |
| dir | City block | 77.43 | 75.68 |
| | Euclidean | 74.63 | 73.04 |
| | Shepard | 40.98 | 41.57 |

### 4.3. Effect of Window Size, Number of Cells and Feature Size

As mentioned in Section 3.3, these three parameters determine the robustness of the object feature and thereby the recognition performance. Figure 3(a) shows the recognition accuracy for the proposed method for different window sizes and different values of X with K = 1. The graph shows the effect of window size and number of cells in recognition performance. The degree of overlap (k) determines whether or not all the pixel values are chosen for feature extraction. In order to analyse their effects, the value of k is fixed as 1 and the values of W and X are varied from 2 to 128. The value of k is fixed at 1 so that all the pixel values are used atleast once for the computation of the feature vector. From the graph, it can be seen that the recognition accuracy is increased as the window size (W) is increased. As mentioned in Section 3.3, the feature vector length will be longer for smaller W. On the other hand, the recognition accuracy gets increased as X increases and later it gets decreased. The recognition accuracy was higher for a value of X=4. So this value is selected as the final value of X. Figure 3(b) shows the variation in recognition accuracy for different values of k. The values of W and X is fixed as 16 and 4 respectively. It is observed that there is only a slight increase in the recognition accuracy as the value of k is increased. So the value of k is selected as 2 so that each pixel value is selected twice for the computation of the feature vector. In all the above experiments, the feature vector pixmagdir with Prewitt gradient was used.

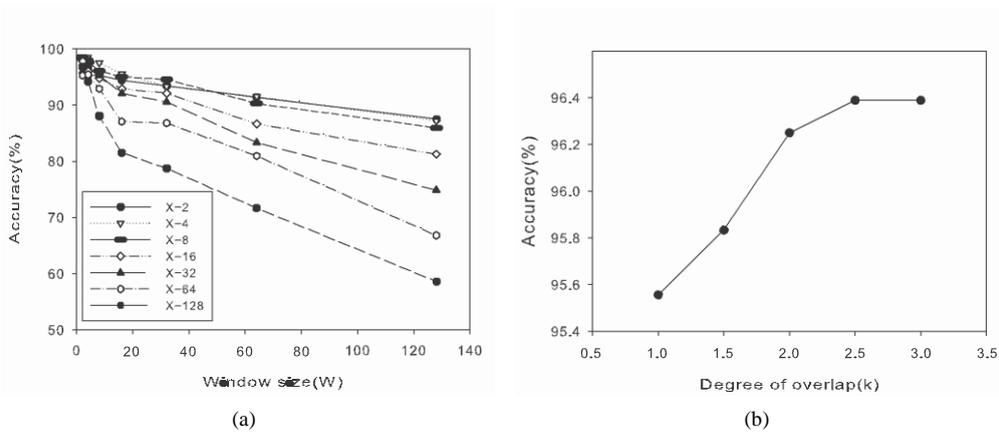

Figure 3: Graph (a) shows the recognition accuracy for various window sizes and number of cells. Graph (b) shows the recognition accuracy for different values of k with W=16 and X = 4. The feature vector consists of pixel values, gradient magnitude and gradient direction.

### 4.4. Noise Robustness

To test the effect of noise on the recognition accuracy, the test images were degraded with noise and the recognition accuracy is computed. The robustness of the proposed method against Gaussian noise, salt and pepper noise and speckle noise are studied. Figures 4(a), and 4(b) shows the recognition accuracy when Gaussian noise and speckle noise of different variances were added to the test images. Figure 4(c) shows the recognition accuracies when salt and pepper noise with different noise density was added to the test images. It is observed that increase in noise levels results in reduced recognition accuracy as a result of increased intra-class variability between the test and training images. The feature vector consists of pixel values, gradient magnitude and gradient direction with parameter values W=16, X=4 and k=2.

### 4.5. Image Scaling and Translation Effects

Figure 5(a) shows the impact of translation on recognition accuracy. Here, the test images were first subjected to pixel translations and the recognition accuracy computed. Shifting is done in the horizontal direction as well as in the vertical directions. From the graph it is observed that the recognition accuracy is least affected by horizontal shifting of the test images. To test the robustness of the proposed technique against scaling effect of images, the test images were scaled and the recognition accuracy computed. It is evident from Figure 5(b) that the proposed method is resilient to minor changes in scaling, while recognition performance drops with increasing variability in scaling.

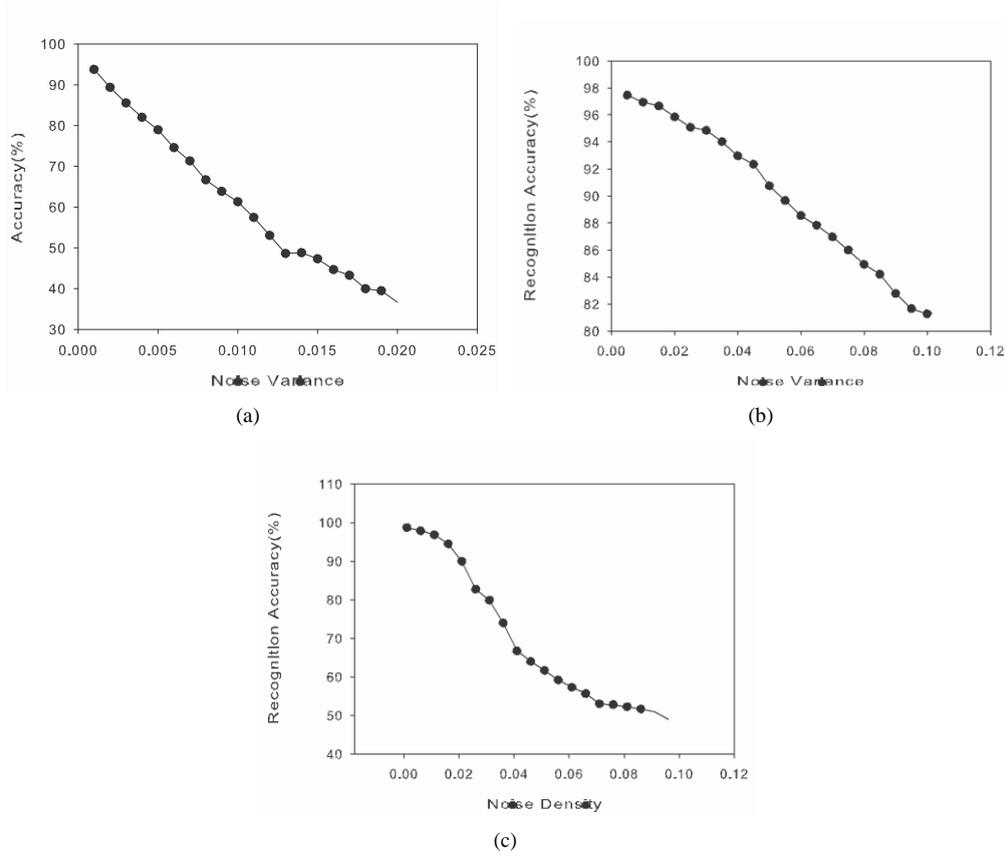

Figure 4: Graphs (a) and (b) shows the recognition accuracies when Gaussian noise and speckle with different noise variances are added to the test images, respectively. Graph (c) shows the recognition accuracies when salt and pepper noise with different noise densities are added to the test images. The feature vector consists of pixel values, gradient magnitude and gradient direction.

4.6. Variation in recognition accuracy with number of classes and training images

Figure 6(a) shows the variation in the recognition accuracy when the number of classes is changed from 200 to 1000. As can be seen from the graph, the recognition accuracy gets increased as the number of classes is reduced. Figure 6(b) shows the plot of recognition accuracy when the number of training images per class is varied. The performance of the system is very much improved with increase in number of training images. The recognition accuracy reaches 99.97% when the number of training images used is 32 which. i.e., when 44% of the images are used for training, a recognition accuracy of 99.97% is achieved. Also the accuracy reached 100% when 88% of images (64 images) are used for training.

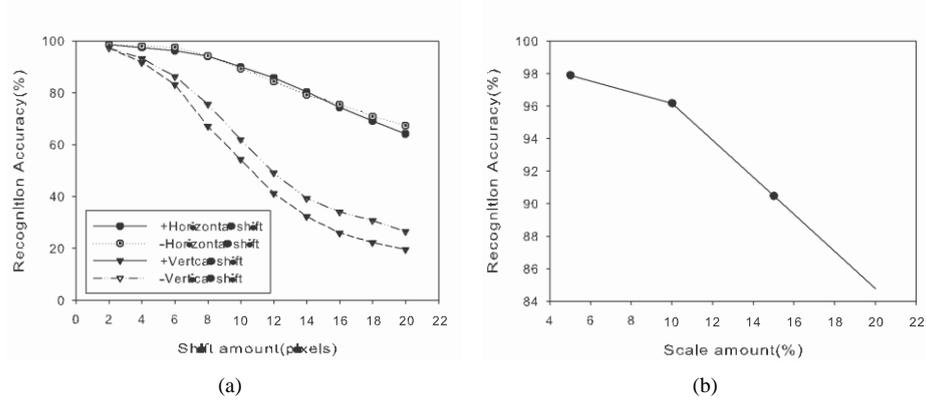

(a)      (b)

Figure 5: Graph (a) shows the change in recognition accuracies when the test images were subjected to translation, i.e., pixel shifts in horizontal and vertical directions. Graph (b) shows the change in recognition accuracy when the test images were subjected to scaling. The feature vector consists of pixel values, gradient magnitude and gradient direction.

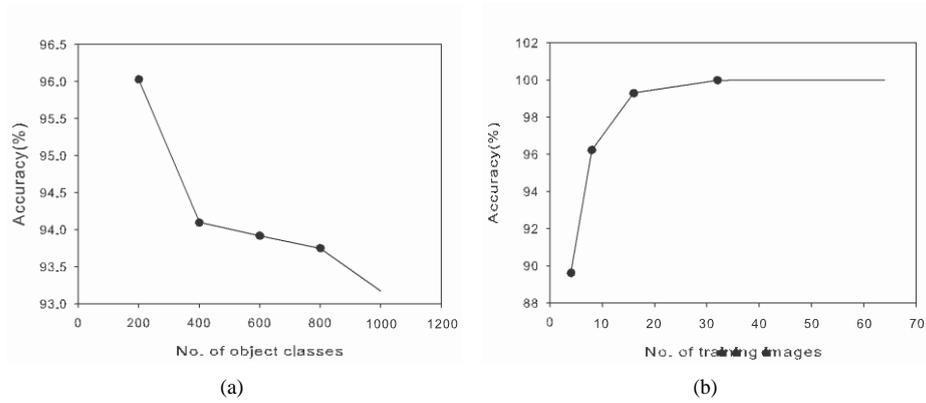

(a)      (b)

Figure 6: Graph (a) shows the change in recognition accuracies when the number of classes is varied. Graph (b) shows the change in recognition accuracy when the number of training images is varied. The feature vector consists of pixel values, gradient magnitude and gradient direction.

Table 2: Comparison of proposed method with existing techniques of object recognition

| Method | Recognition Accuracy(%) | |
|---|---|---|
| | ALOI | COIL-100 |
| Proposed method | 93.18 | 92.39 |
| Gabor Jet based object representations[48] | 83 | - |
| SalBayes[19] | 83.83 | - |
| SIFT[19] | 72.68 | 87.19 |
| HMAX[19] | 83.42 | 77.02 |
| SNoW/intensity[41] | - | 85.13 |
| SNoW/edges[41] | - | 89.23 |
| Linear SVM[41] | - | 84.80 |
| Nearest Neighbour[41] | - | 79.52 |

Table 3: Performance of proposed method on Pascal visual object challenge 2007 database

|  | Proposed method | INRIA Genetic[45] | INRIA Flat[53] | XRCE[37] | QMUL LSPCH[53] | TKK[46] |
|---|---|---|---|---|---|---|
| Aeroplane | 100.0 | 77.5 | 74.8 | 72.3 | 71.6 | 71.4 |
| Bicycle | 98.7 | 63.6 | 62.5 | 57.5 | 55.0 | 51.7 |
| Bird | 81.1 | 56.1 | 51.2 | 53.2 | 41.1 | 48.5 |
| Boat | 100.0 | 71.9 | 69.4 | 68.9 | 65.5 | 63.4 |
| Bottle | 47.5 | 33.1 | 29.2 | 28.5 | 27.2 | 27.3 |
| Bus | 96.9 | 60.6 | 60.4 | 57.5 | 51.1 | 49.9 |
| Car | 94.1 | 78.0 | 76.3 | 75.4 | 72.2 | 70.1 |
| Cat | 24.0 | 58.8 | 57.6 | 50.3 | 55.1 | 51.2 |
| Chair | 16.4 | 53.5 | 53.1 | 52.2 | 47.4 | 51.7 |
| Cow | 81.7 | 42.6 | 41.1 | 39.0 | 35.9 | 32.3 |
| Diningtable | 88.6 | 54.9 | 54.0 | 46.8 | 37.4 | 46.3 |
| Dog | 23.6 | 45.8 | 42.8 | 45.3 | 41.5 | 41.5 |
| Horse | 88.7 | 77.5 | 76.5 | 75.7 | 71.5 | 72.6 |
| Motorbike | 99.1 | 64.0 | 62.3 | 58.5 | 57.9 | 60.2 |
| Person | 18.5 | 85.9 | 84.5 | 84.0 | 80.8 | 82.2 |
| Pottedplant | 24.2 | 36.3 | 35.3 | 32.6 | 15.6 | 31.7 |
| Sheep | 64.5 | 44.7 | 41.3 | 39.7 | 33.3 | 30.1 |
| Sofa | 34.4 | 50.6 | 50.1 | 50.9 | 41.9 | 39.2 |
| Train | 100.0 | 79.2 | 77.6 | 75.1 | 76.5 | 71.1 |
| TVmonitor | 98.7 | 53.2 | 49.3 | 49.5 | 45.9 | 41.0 |
| Average Accuracy | 69.0 | 59.4 | 57.5 | 55.6 | 51.2 | 51.7 |

4.7. Comparison with Other Methods

Table 2 shows the comparison between the proposed method and some of the commonly used recognition algorithms for object recognition on the ALOI and COIL-100 databases [48][19][41]. It is evident from the table that the proposed object recognition algorithm improved the recognition accuracy by 10% in ALOI and 8% in case of COIL-100 databases compared to previous object recognition algorithms. Table 3 compares the performance of the proposed method on the Pascal visual object challenge 2007 database [20] with several other algorithms [45][53][37][53][46]. The parameters used are W=16, X=4, k=2 with Prewitt mask as gradient filter. Thus the proposed method outperformed the previous object recognition methods in terms of object recognition accuracy. There are several reasons for this improved recognition accuracy. The distributed pixel selection technique in the feature extraction provides robustness against local variations in an image. Local variations in an image may be defined as changes that occur to small regions of the image. In the pixel selection step, pixels from random positions are selected, added together and stored in cells. Since the pixels are selected in a distributed manner, the local variations also gets distributed among the cells and their effect on the feature vector gets nullified. The other type of variation that can occur in images is global variations that causes a change in pixel values throughout the image such as those resulting from scaling or shifting of images, and illumination changes. The inhibition step in Fig. 1(c) helps in minimising the global effect on the images. In the inhibition step, the cells are grouped together and in each group, the cell with the highest value is changed to 1 and all others are changed to 0. If a global variation occur, then the all the values in the cells will get increased. But the inhibition step will eliminate this effect, since all the cells get increased with almost the same value. Selection of distances measures such as Shepard's similarity measure also has an important role in the increased recognition accuracy. For example, the Shepard's similarity measure is known to eliminate the effect of outliers present in the feature vectors thereby reducing the intra-class mismatches and improving the recognition accuracy.

5. Conclusion

We presented a set of sparsely distributed selection of pixels for a fused encoding scheme using both object image pixel values and their pixel gradients. The properties of hierarchy, modularity, and sparsity constraints inspired from human visual processing system were applied to develop the proposed object recognition system. Hierarchy was implemented using multi-levels of processing that involved the feature grouping and bit plane processing followed by feature encoding. Modularity was incorporated by arranging and randomly selecting features in groups and repeating the same process over several times to form multiple feature groups. The sparsity constraints are imposed by the random selection of features to the groups and binarization within the group that can result in the exclusion of least important information from the original image. We demonstrated improved recognition accuracies for the proposed method when tested on the Amsterdam library of object images (ALOI) database, Columbia object image library (COIL-100) database and PASCAL visual object challenge 2007 database, validating the claims of hierarchy, modular processing and sparsity constraints being a useful concepts for implementing automated object recognition. The robustness of the proposed method against variability in noise, scaling and horizontal shifts was tested. The comparison with 8 different methods indicated a statistically significant improvement in recognition accuracies. In addition, the proposed method due to its modular processing ability supports implementation in scalable parallel processing architectures. The results indicate that like most of object matching methods the proposed method is not scale and shift invariant for larger variations between the test and train images. In addition, the noise also would have an impact in the recognition accuracies, as the linear difference operations in the gradient filtering improves the feature diversity, however, cannot inhibit noise. This remains a challenge and is a necessary future work to improve the robustness of the method.